%% file: main.tex
\RequirePackage{snapshot}
\documentclass[letterpaper]{article}
\usepackage{proceed2e}
\usepackage[margin=1in]{geometry}
\usepackage{mycustom}
\usepackage{subcaption}

\usepackage{times}

\title{Uncertainty in the Variational Information Bottleneck}

\author{  {\bf Alexander A.~Alemi}\\
\texttt{alemi@google.com} \\
Google Research \\
\And
Ian Fischer \\
\texttt{iansf@google.com} \\
Google Research \\
\And
Joshua V. Dillon \\
\texttt{jvdillon@google.com} \\
Google Research}

\begin{document}

\maketitle

	\InputIfFileExists{abstract}{}{\typeout{No file abstract.}}%

	\InputIfFileExists{intro}{}{\typeout{No file intro.}}%

	\InputIfFileExists{vib}{}{\typeout{No file vib.}}%

	\InputIfFileExists{experiments}{}{\typeout{No file experiments.}}%

	\InputIfFileExists{conclusion}{}{\typeout{No file conclusion.}}%

\clearpage



\bibliographystyle{icml2018}
\bibliography{bib}

\newpage
\appendix
	\InputIfFileExists{appendix}{}{\typeout{No file appendix.}}%

\end{document}

%% file: abstract.tex
\begin{abstract}

We present a simple case study, demonstrating that Variational Information Bottleneck (VIB)
can improve a
network's classification calibration as well as its ability to detect
out-of-distribution data.
Without explicitly being designed to do so, VIB gives two natural metrics for
handling and quantifying uncertainty.

\end{abstract}

%% file: intro.tex
\section{Introduction}


It is important to track and quantify uncertainty.
Prediction uncertainty is a consequence of one or more non-exclusive
sources including \citep{uncertainty_deep_learning}, but not limited to:
\emph{aleatoric uncertainty}, (e.g., noisy data or measurement imprecision)
\emph{epistemic uncertainty}, (e.g., unknown model parameters, unknown model structure)
and \emph{out-of-distribution samples} (e.g., train/eval datasets do not share the same stochastic generator).

Ideally a model should have some sense of whether it has sufficient evidence
to render a prediction. Adversarial examples~\citep{adversaries} demonstrate a
broad failure of existing neural networks to be sensitive to out-of-distribution
shifts, but distributional shifts don't require an adversary--any deployment of a model
in a real-world system is likely to encounter surprising out-of-distribution data.

Most classifiers report probabilities.
Models which do this \emph{well} are said to be
\emph{calibrated} -- the observed occurrence frequency matches the
predicted frequency~\citep{calibration}. 
If the model assigns a
20\% probability to an event, that event should occur 20\% of the time.
Current deep networks tend to be poorly calibrated~\citep{oncalibration},
making highly confident predictions even when they are wrong.

Many recent
papers~\citep{oncalibration, ensemble, baseline, odin, confidence,
learningconfidence} have proposed techniques for improving the quantification
of uncertainty in neural networks.
The simplest such modification, \emph{temperature
scaling}~\citep{oncalibration,baseline,odin}, changes the temperature of
the softmax classifier after training for use during prediction. It empirically
does well at improving the calibration and out-of-distribution detection
for otherwise unmodified networks.  Other approaches require larger
interventions or modifications, e.g. training on out-of-distribution
data directly, generated either in an adversarial
fashion~\citep{odin,confidence} or chosen by the
practitioner~\citep{learningconfidence}.

Instead of specifically trying to invent a new technique for improved
calibration, this work empirically demonstrates that the previously described
\emph{variational information bottleneck} (VIB)~\citep{vib} gives calibrated predictions
and does a decent job at out-of-distribution detection
without sacrificing accuracy.

%% file: vib.tex
\section{Variational Information Bottleneck}

Variational Information Bottleneck (VIB)~\citep{vib} learns a variational
bound of the Information Bottleneck (IB)~\citep{infobottle}. VIB is to
\emph{supervised learning} what $\beta$-VAE~\citep{betavae} is to
\emph{unsupervised learning}; both are justified by information theoretic
arguments~\citep{brokenelbo}.

IB regards supervised learning as a representation learning problem,
seeking a 
stochastic map from input data $X$ to some latent representation $Z$ that
can still be used to predict the labels $Y$, under a constraint on its
total complexity.

Writing mutual information between random variables $Z,Y$ as $I(Z;Y)$, the
information bottleneck procedure is:
\begin{equation}
\max I(Z;Y) \text{ subject to } I(Z;X) \le R
\end{equation}
where $R$ is a constant \emph{bottleneck}. Rewritten as an
unconstrained Lagrangian optimization, the procedure is:
\begin{equation}
	\max I(Z;Y) - \beta I(Z;X)
\end{equation}
where $\beta$ controls the size of the bottleneck.

IB is intractable in general. However, there exists
a simple tractable variational bound~\citep{vib}:
\begin{equation}
	\max_{\theta,\phi,\psi} \mathbb{E}_{p(x,y)e_\theta(z|x)}\left[ \log q_\psi(y|z) - \beta \log \frac{e_\theta(z|x)}{m_\phi(z)} \right]
\end{equation}
The $q$ term measures classification log likelihood and the $\frac{e}{m}$ term represents \emph{rate},
penalizing lengthy encodings $e$ relative to some code space, $m$.
More precisely:
\begin{itemize}
  \item $e_\theta(z|x)$ is a learned stochastic \emph{encoder} that transforms the input $X$ to some encoding $Z$;
  \item $q_\psi(y|z)$ is a variational classifier, or \emph{decoder} that predicts the labels $Y$ from the codes;
  \item $m_\phi(z)$ is a variational \emph{marginal} that assigns a density in the code space; and
  \item $p(x,y)$ is the empirical data distribution.
\end{itemize}

VIB incorporates uncertainty in two ways.
VIB is doubly stochastic in the sense that both the underlying feature representation ($Z$)
\textit{and} the labels ($Y$) are regarded as random variables.
Conversely, DNNs only regard the labels as being random variables.
By explicitly modeling the representation distribution,
VIB has the ability to model both mean and variance in the label predictions.
Recall that for most DNNs, the output layer
corresponds to a distribution in which variance is a function of mean 
(e.g., a
binary classifier predicting $p$ for a class occurrence must also predict the
variance $p(1-p)$).\footnote{Of course, predicting a variance of $p(1-p)$ is
reasonable if the model is well-specified, but it almost certainly isn't.}
The stochasticity in the representation induces an effective ensemble of decoder predictions.
Ensembles of whole neural networks have been shown to be well calibrated~\citep{ensemble}.

The second source of uncertainty is provided by the per-instance rates.  Recall that the
rate is the KL divergence between the conditional distribution over codes given
the input and the code space defined by the learned marginal $m_\phi(z)$; i.e.
$\operatorname{KL}[e_{\theta}(z|x)||m_{\phi}(z)]$.
Here, the marginal effectively learns a density model for the data, albeit in the lower-dimensional, 
lower-information code-space rather than the original input space.
Density estimation, whether explicit~\citep{learningconfidence} or
implicit~\citep{gans}, has been shown to be useful for out-of-distribution detection.

A stochastic representation requires computing an additional expectation at both
training and test time. 
We find that approximating the expectation with a Monte Carlo average
with a few dozen samples produces surprisingly low variance predictions.
In practice, the bulk of the model complexity is in the encoding network
(typically a deep convolutional network), which
need only be run once to obtain the parameters for the encoding distribution.
We find that very low dimensional distributions (e.g. 3D fully covariant Gaussians)
and simple variational decoders (a mixture of softmaxes~\citep{softmax}) work well.
For these, sampling and evaluating the log likelihoods 
is dwarfed by the initial encoder, and so VIB has little effect on computation
time.

%% file: experiments.tex
\section{Experiments}

Below we demonstrate results of training a VIB classifier on
FashionMNIST~\citep{fashion}. We compare the network's ability
to both quantify the uncertainty in its own predictions on the test
set, as well as identify when shown out-of-distribution data.

For our encoder we used a 7-2-WRN~\citep{wrn} initialized with the delta-orthogonal
initializer~\citep{deltaorthogonal} and bipolar-relu non-linearities~\citep{bipolar}
topped with a 3-dimensional
fully covariant Gaussian encoding distribution. Given the strong
spatial inhomogeneities of the data, we concatenated a $28 \times 28 \times 5$
set of learned parameters to the original image before feeding it through the
network.
The variational marginal
is a mixture of 200 3-dimensional fully covariant Gaussians. The decoder
is a Categorical mixture of five Categorical distributions with affine logits.
For the baseline deterministic network, the 7-2-WRN feeds directly into
a logistic classifier~\footnote{
In \citet{softmax}, the authors argue for using a Mixture of Softmaxes if the model's
representation has lower dimensionality than the number of output classes.
This is the case for the VIB model with a 3D latent space, but not for the classifier,
which has 128 dimensions coming from the encoder.
}.
The networks were trained with a 0.1 dev-decay~\citep{devdecay}
with Adam~\citep{adam}, with initial learning rate of 0.001. All other hyperparameters
are set to the TensorFlow defaults.
We emphasize that we used no form of regularization aside from VIB.

\begin{figure}[ht!]
	\centering
	\includegraphics[width=1.0\linewidth]{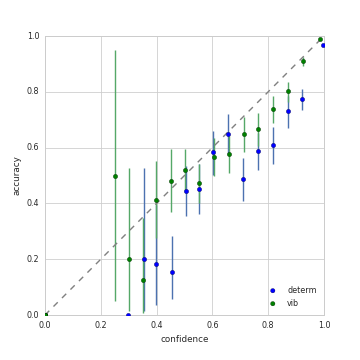}
	\caption{\label{fig:calibration}Reliability diagram for the trained networks.
	This shows how well-calibrated the networks are.  The predictions were split into 20
  equally-sized bins.  The accuracy was measured in each bin.  Shown is the accuracy as well as the 90\% confidence
	interval for the accuracies. A perfectly calibrated model would fall on the diagonal.
	}
\end{figure}

\def \supwidth {0.45}

\begin{figure}[ht!]
   \centering
  \begin{subfigure}[t]{\supwidth\linewidth}
    \includegraphics[width=\linewidth]{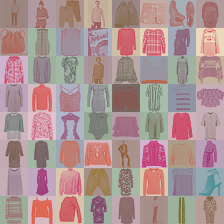}
    \caption{\label{fig:lowp}Lowest classification confidence examples.  }
  \end{subfigure}
  \begin{subfigure}[t]{\supwidth\linewidth}
      \includegraphics[width=\linewidth]{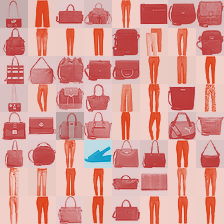}
    \caption{\label{fig:highp}Highest classification confidence examples.  }
  \end{subfigure}

  \begin{subfigure}[t]{\supwidth\linewidth}
    \includegraphics[width=\linewidth]{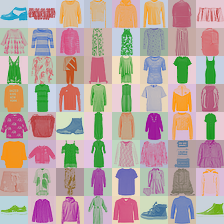}
    \caption{\label{fig:misclass}Highest confidence misclassification examples.}
  \end{subfigure}
  \begin{subfigure}[t]{\supwidth\linewidth}
    \includegraphics[width=\linewidth]{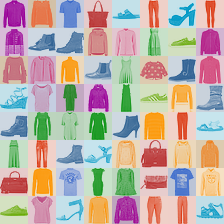}
    \caption{\label{fig:random}Random selection.}
  \end{subfigure}

  \begin{subfigure}[t]{\linewidth}
  \includegraphics[width=\linewidth]{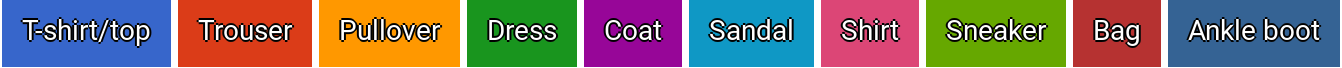}
  \caption{\label{fig:legend}Color Legend for the FashionMNIST classes.}
  \end{subfigure}

     \caption{\label{fig:examples}Extreme examples from the test set. Images
     from the test set are shown colored according to a weighted combination of the class
     colors in \Cref{fig:legend}.  The background
     lightness shows $R$, and the background hue is the correct class.
     }
\end{figure}


%
	\InputIfFileExists{correcttable}{}{\typeout{No file correcttable.}}%

\InputIfFileExists{auctable}{}{\typeout{No file auctable.}}%

\paragraph{Metrics.}
We can compare the ability of various signals to distinguish misclassified examples
beyond the reported probability ($p(y|x)$). For instance,
we can look at the entropy of the classifier $H(y|x) = -\sum_i p(y_i|x) \log
p(y_i|x)$ ($H$ in the tables); or in the case of VIB networks, the rate ($R$ in the tables) and
$p(y,z|x) = \int dx\, p(x) e_{\theta}(z|x) q_{\psi}(y|z) \approx m_{\phi}(z) q_{\psi}(y|z)$,
the conditional joint likelihood.
See~\Cref{sec:appendix} for results and discussion of all four metrics. Here we focus on $H$ and $R$.

Since these are not necessarily calibrated measures themselves, we compute threshold independent metrics
(AUROC, AUPR) as done in \citet{baseline,odin} (See references for detailed
definitions of metrics). In \Cref{tab:correct,tab:ood}, \textit{FPR} is the false positive rate,
\textit{TPR} is the true positive rate, \textit{AUROC} is the area under the
ROC curve, and \textit{AUPR} is the area under the precision-recall curve.
AUPR is sensitive to which class is denoted as positive, so both
versions are reported. \textit{AUPR In} sets the in-distribution examples as positive.
\textit{AUPR Out} sets the out-of-distribution examples as positive.

\paragraph{Discussion.}
In \Cref{fig:calibration} we demonstrate that the VIB network is better-calibrated
than the baseline deterministic classifier. The deterministic classifier
is overconfident.

We report error detection results in \Cref{tab:correct}. Note that the deterministic
baseline and VIB perform equivalently well across the board.
\textit{Baseline T} in the tables is the baseline model tested at a temperature of 100.
$H$ seems to offer better error detection than $R$ for the VIB network.

Next we measure the ability to detect out-of-distribution data. We take the
same networks, trained on FashionMNIST, and then evaluate them on the
combination of the original FashionMNIST test set, as well as another test set.
We evaluate the ability of each metric
to distinguish between the two.
We compare against randomly generated images (U(0,1), for uniform noise) and MNIST digits to get a
measure of gross distribution shift. We also test on horizontally
and vertically flipped FashionMNIST images for more subtle distributional shift.
We believe these offer an interesting test since the FashionMNIST data has a strong orientation --
the clothes have clear tops versus bottoms, and for the shoes, care was taken to
try to have all the shoes aligned to the left. Since these modified image distributions
are just mirrored versions of the original, all of the mirror invariant statistics of the images
are unchanged by this operation, suggesting this is a more difficult situation to resolve than
the first two test sets.

From \Cref{tab:ood} we can draw a few early conclusions.
Temperature scaling remains a powerful post-hoc method for improving the calibration and
out-of-distribution detection of networks.
It additionally improved the performance of our VIB networks (not shown here),
but here we emphasize that VIB networks offer an improvement
over the baseline without post-hoc modification.
Without relying on temperature scaling, a VIB network can
use $R$ for stronger out-of-distribution detection.



To visually demonstrate the different measures of uncertainty in the VIB
network, in \Cref{fig:examples} we show some of
the extreme inputs.
Interestingly, the network has instances where the rate is low
(signifying that it thinks the example is valid clothing)
but that have high predictive entropy (signifying that the network doesn't know how to classify them)
as well as the converse--things that look outside the training distribution, but the network
is certain they are of a particular class.
E.g., in \Cref{fig:lowp} there are some examples with both low rate and low classification confidence.
These examples are within the data distribution as far as the network is concerned,
but it is uncertain about their class, often splitting its prediction equally across two options.
In \Cref{fig:highp} we see the opposite, with some examples like the first having very high
classification confidence (the network is certain that is a handbag),
but with high rate (it is unlike most other handbags, given its unusually long handle).
This sort of distinction is not possible in ordinary networks.
\Cref{fig:misclass} shows the most confidently misclassified examples from the test set.
Arguably, a majority of these instances are mislabelings in the test set itself.
\Cref{fig:random} shows a representative sample of images from the test set for comparison.
\Cref{fig:fashion,fig:vfashion,fig:hfashion} in~\Cref{sec:appendix} show the complete test set, as well as its mirrored
versions.


%% file: correcttable.tex
\begin{table*}[htb]
	\centering
	\small
	\begin{tabular}{|c|c|c|c|c|c|c|}
		\hline
		Method &  \makecell{Accuracy} & Threshold & \makecell{FPR @95\% TPR $\downarrow$} & AUROC $\uparrow$ & \makecell{AUPR In $\uparrow$} & \makecell{AUPR Out $\uparrow$} \\
		\hhline{|=|=|=|=|=|=|=|}

		Baseline & \multirow{2}{*}{\textbf{92.9}} & \multirow{2}{*}{$H$} & 45.3 & \textbf{91.3} & \textbf{99.3} & 03.8 \\
		\cline{0-0}
		\cline{4-7}
		Baseline T &  &  &  83.0 & 74.7 & 97.6 & \textbf{04.3} \\


		\hhline{|=|=|=|=|=|=|=|}

		\multirow{2}{*}{VIB} & \multirow{2}{*}{\textbf{\textit{92.8}}} & $H$ & \textbf{44.8} & \textbf{\textit{91.2}} & \textbf{\textit{99.2}} & 03.8 \\
		\cline{3-7}
		& & $R$ & 63.1 & 84.7 & 98.6 & \textbf{\textit{03.9}} \\

		\hline


	\end{tabular}

	\caption{\label{tab:correct}Results on error detection. Here we use the threshold signal to try to distinguish correct and incorrect predictions. Arrows denote whether higher or lower scores are better.
	\textbf{Bold} indicates the best score in that column.
	\textbf{\textit{Bold italics}} indicate the closest VIB score if a baseline score was better.
	Baseline T is with temperature scaling.
	VIB mostly matches performance. For VIB, $H$ is better at error detection than $R$, while the opposite is true with OoD detection.
	}
\end{table*}

%% file: auctable.tex
\begin{table*}[htb]
	\centering
	\small
	\begin{tabular}{|c|c|c|c|c|c|c|}
	\hline
		OoD & Method & Threshold & \makecell{FPR @\\95\% TPR $\downarrow$} & \makecell{AUROC \\ $\uparrow$} & \makecell{AUPR \\ In $\uparrow$} & \makecell{AUPR \\ Out $\uparrow$} \\
		\hhline{|=|=|=|=|=|=|=|}

	\multirow{4}{*}{U(0,1)}

		& Baseline & \multirow{2}{*}{$H$} &  79.5 & 87.3 & 91.2 & \textbf{32.9} \\
		\cline{2-2}
		\cline{4-7}
		& Baseline T &  & \textbf{00.6} & \textbf{99.2} & \textbf{99.5} & 30.7 \\
		\cline{2-2}
		\cline{3-7}
		& \multirow{2}{*}{VIB} & $H$ & \textbf{\textit{43.8}}  & \textbf{\textit{90.8}} & 91.0 & 31.8 \\
		& & $R$ & \textbf{\textit{09.6}}  & \textbf{\textit{97.6}} & \textbf{\textit{98.0}} & 30.9 \\


	\hhline{|=|=|=|=|=|=|=|}

	\multirow{4}{*}{MNIST}

		& Baseline & \multirow{2}{*}{$H$} &  78.4 & 73.2 & 75.4 & \textbf{36.1} \\
		\cline{2-2}
		\cline{4-7}
		& Baseline T &  &  41.9 & 88.9 & 90.8 & 31.8 \\
		\cline{2-2}
		\cline{3-7}
		& \multirow{2}{*}{VIB} & $H$ & \textbf{\textit{63.3}}  & \textbf{\textit{83.2}} & \textbf{\textit{83.2}} & 33.4 \\
		& & $R$ & \textbf{27.3}  & \textbf{92.7} & \textbf{92.0} & 31.4 \\


	\hhline{|=|=|=|=|=|=|=|}

	\multirow{4}{*}{HFlip}

		& Baseline & \multirow{2}{*}{$H$} &  86.7 & 66.0 & 64.4 & 39.8 \\
		\cline{2-2}
		\cline{4-7}
		& Baseline T & & 77.1 & 67.8 & \textbf{64.7} & 39.3 \\
		\cline{2-2}
		\cline{3-7}
		& \multirow{2}{*}{VIB} & $H$ & \textbf{\textit{81.1}} & 65.7 & 59.8 & \textbf{42.3} \\
		& & $R$ & \textbf{73.0}  & \textbf{68.7} & \textbf{\textit{64.4}} & 39.3 \\


	\hhline{|=|=|=|=|=|=|=|}

	\multirow{4}{*}{VFlip}

		& Baseline & \multirow{2}{*}{$H$} &  64.5 & 85.6 & 87.8 & \textbf{32.8} \\
		\cline{2-2}
		\cline{4-7}
		& Baseline T & &  38.5 & \textbf{92.1} & \textbf{94.1} & 31.5 \\
		\cline{2-2}
		\cline{3-7}
		& \multirow{2}{*}{VIB} & $H$ & \textbf{\textit{45.0}}  & \textbf{\textit{86.4}} & 83.4 & 32.6 \\
		& & $R$ & \textbf{34.4}  & \textbf{\textit{89.7}} & 87.0 & 32.0 \\

		\hline
	\end{tabular}

	\caption{\label{tab:ood}Results for out-of-distribution detection (\textit{OoD}). Arrows denote whether higher or lower scores are better.
	\textbf{Bold} indicates the best score in that column for a particular out-of-distribution dataset.
	\textbf{\textit{Bold italics}} indicate VIB scores that are better than the \textit{Baseline} row.
	}
\end{table*}

%% file: conclusion.tex
\section{Conclusion}

While more experimentation is needed, initial investigations suggest that VIB
produces well-calibrated networks, while also giving quantitative measures
for detecting out-of-distribution samples. 
Coupled with VIB's other demonstrated
abilities to improve generalization, robustness to adversarial examples, preventing memorization
of random labels, and becoming robust to nuisance variables~\citep{vib,achille}, we 
believe it deserves more investigation experimentally and theoretically.



%% file: appendix.tex
\section{Appendix}
\label{sec:appendix}

Here we show more detailed images and tables for our results.

\begin{figure}[htb]
	\centering
	\includegraphics[width=\linewidth]{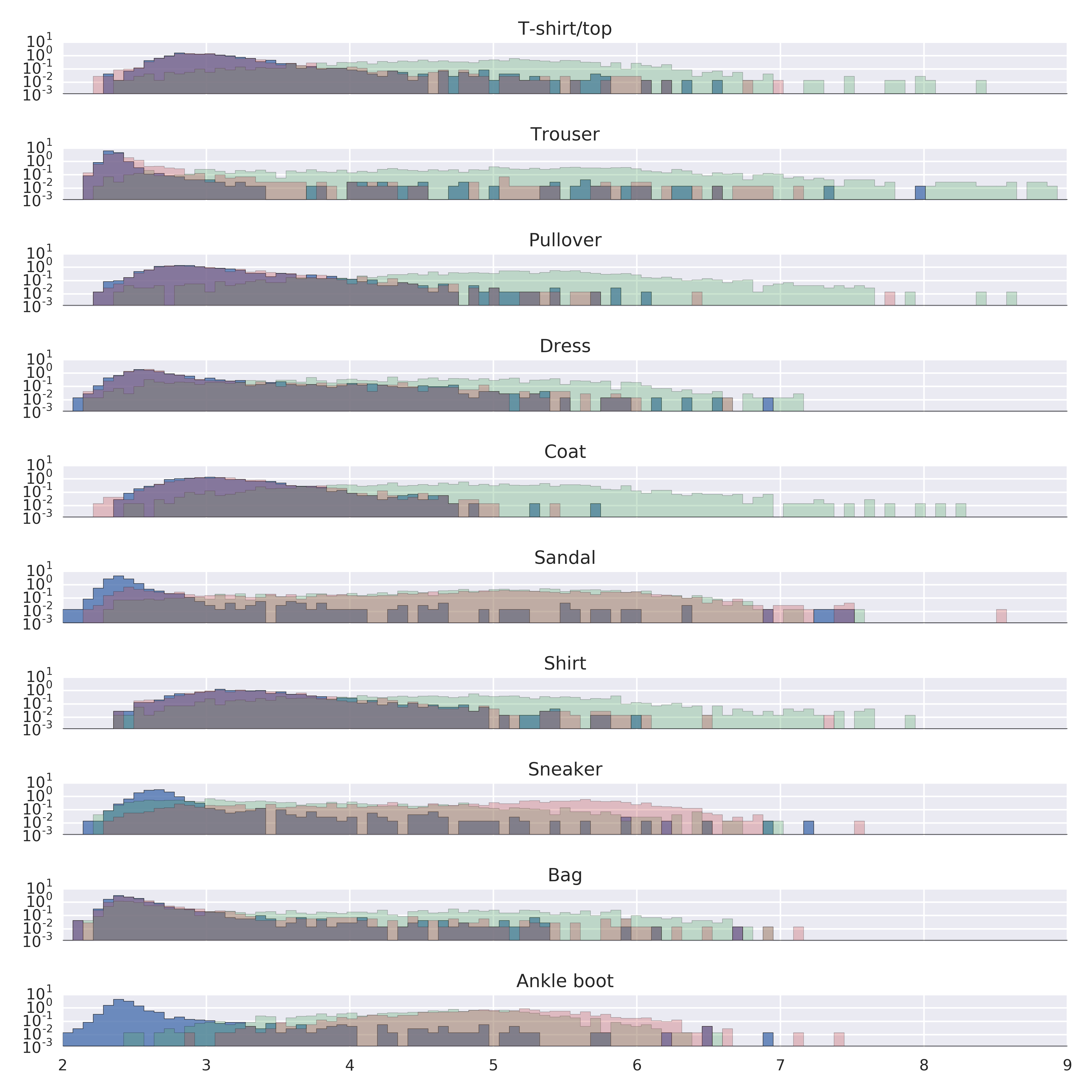}
	\caption{\label{fig:rate-breakdown}
	Demonstration of the change in rates when we perturb the images. The
	original test set histogram is in blue, the vertically flipped images
	are in green, and the horizontally flipped images are in red.  Notice
	that this is a log-histogram, and there is very little overlap between
	the original images and the vertically flipped ones in any class.  For
	most classes there is a great deal of overlap between the original
	images and the horizontally flipped ones, except for Sandal, Sneaker,
	and Ankle Boot, which have a strong left-right asymmetry.
	}
\end{figure}

\Cref{fig:rate-breakdown} shows how the rate responds as we flip the images
vertically and horizontally on a per-class basis. It demonstrates the model has learned
useful semantics regarding the orientation of the clothes categories.

\begin{figure}[htb]
	\centering
	\includegraphics[width=\linewidth]{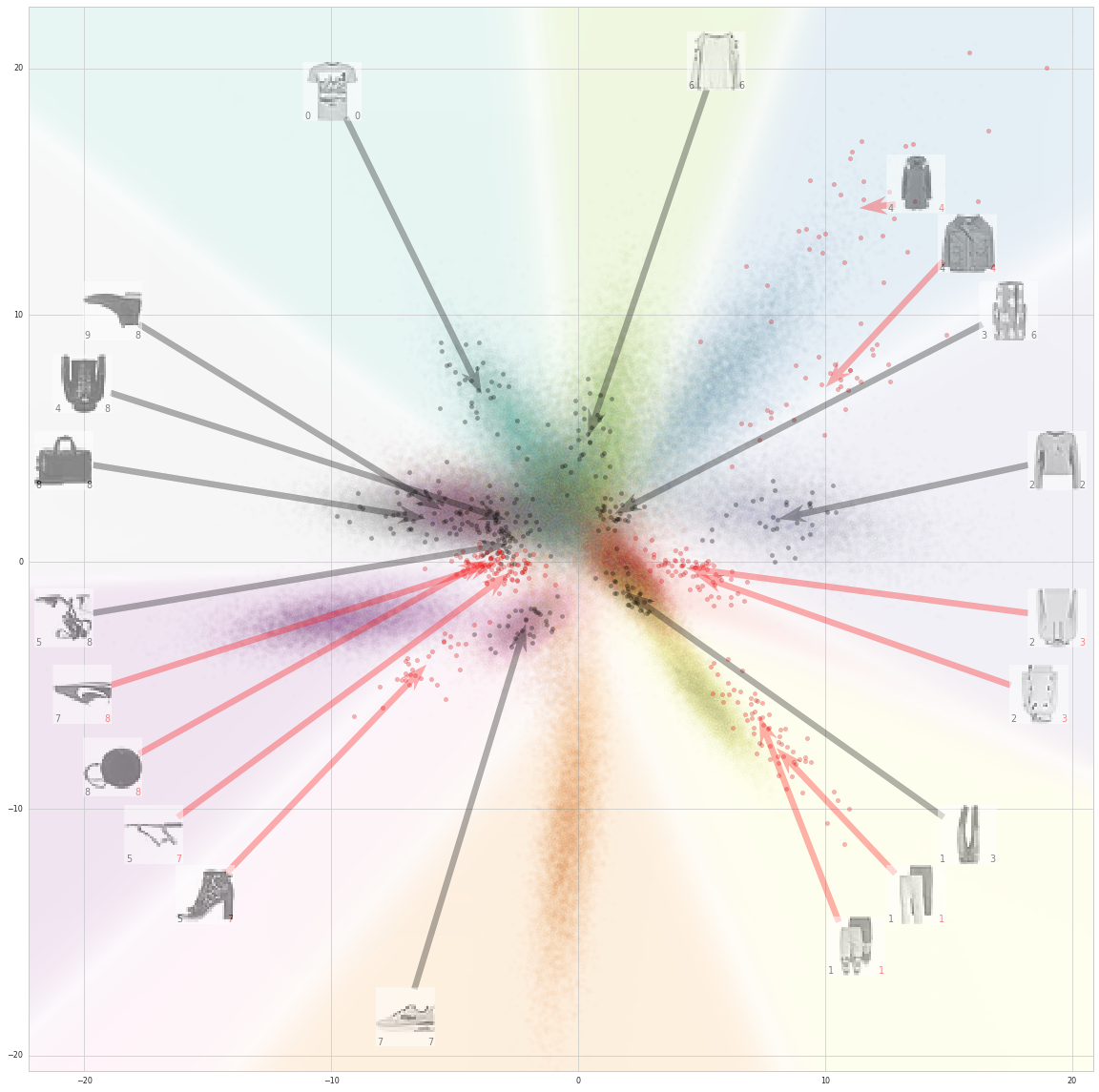}
	\caption{\label{fig:2dlatent}
	Visualization of a VIB model's 2D latent space.  The background color
	corresponds to the mixture of classes predicted by the classifier for
	that location in $z$ space.  Transparency increases as the confidence
	of the maximum class decreases, so the light white lines in the image
	correspond to a high $H(Y|Z)$. More saturated regions correspond to
	higher marginal densities, $m(z)$. The colored points are the $z$
	locations for 20,000 test set images -- 10,000 normal images, and
	10,000 vertically-flipped images. The color of each point indicates its
	true label, and should match the color of the background.
	Additionally, 10 clouds of dark grey points are randomly selected
	images from the test set, 1 for each class. Each cloud corresponds to
	the 32 samples from the encoder taken for each input. The clouds show
	the variance in the encoding distribution.  Images with dark grey
	arrows point to the mean of the corresponding cloud of 32 samples.
	Finally, 10 clouds of dark red points are the 10 images from the test
	set with the highest mean rate, $R$. Similarly, the images with dark
	red arrows point to the mean of the corresponding cloud of 32 encoder
	samples.
	}
\end{figure}

\Cref{fig:2dlatent} shows a visualization of the latent space of a VIB model with the same
architecture as the model described in the paper, but with a 2D latent space rather than 3D.
In general we can train accurate classifiers with very low dimensional representations.
Note that the images with red arrows, which are the 10 highest rate ($R$) examples in the
test set, occur either near classification boundaries, or far away from the high density regions.
However, they are not necessarily incorrectly classified. For example, all three images in the lower
right are images of pants that are correctly classified. The two images with red arrows are images
that contain two pairs of pants, which is unusual in the dataset.
Similarly, the highest rate images are not all out-of-distribution as defined by
vertical flip. In fact, only 5 of the top 10 highest rate images are vertically flipped
in this model with this set of samples. Of the remaining 5, two are the previously mentioned
pants, two are ``unusual'' coats according to the model, and the final one has a true
label of ``sandal'', but is mislabeled by the model as ``sneaker''.

\begin{figure*}[htb]
	\centering
	\includegraphics[width=\textwidth]{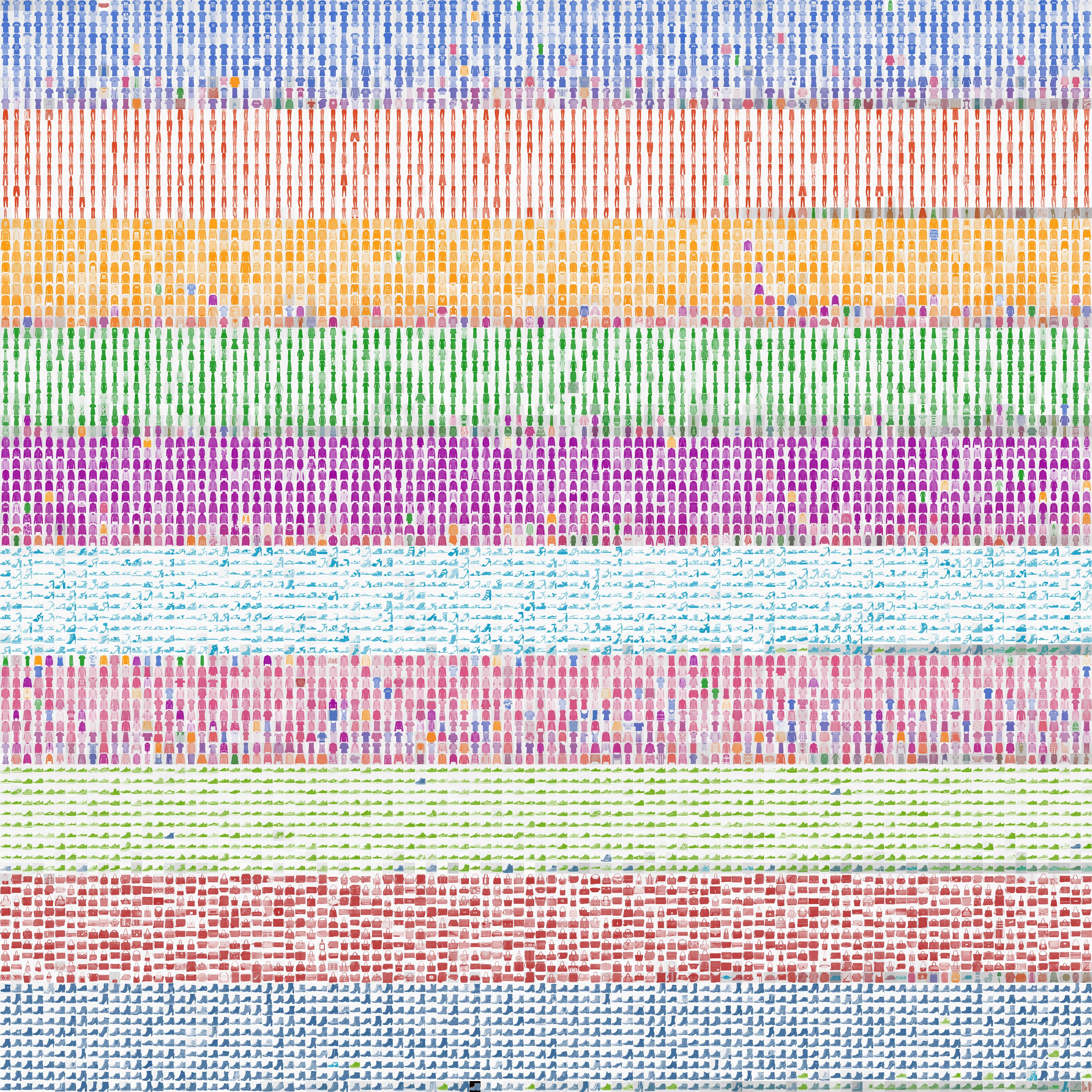}
	\caption{\label{fig:fashion}
	Fashion MNIST. The complete test set.  Foreground color is the weighted
	mixture of the predicted class colors.  Background darkness is the per-instance
	rate.  The images are ordered first by the true class into blocks, and then in each block 
	by maximum $p(y|x)$.
	}
\end{figure*}

\begin{figure*}[htb]
	\centering
	\includegraphics[width=\textwidth]{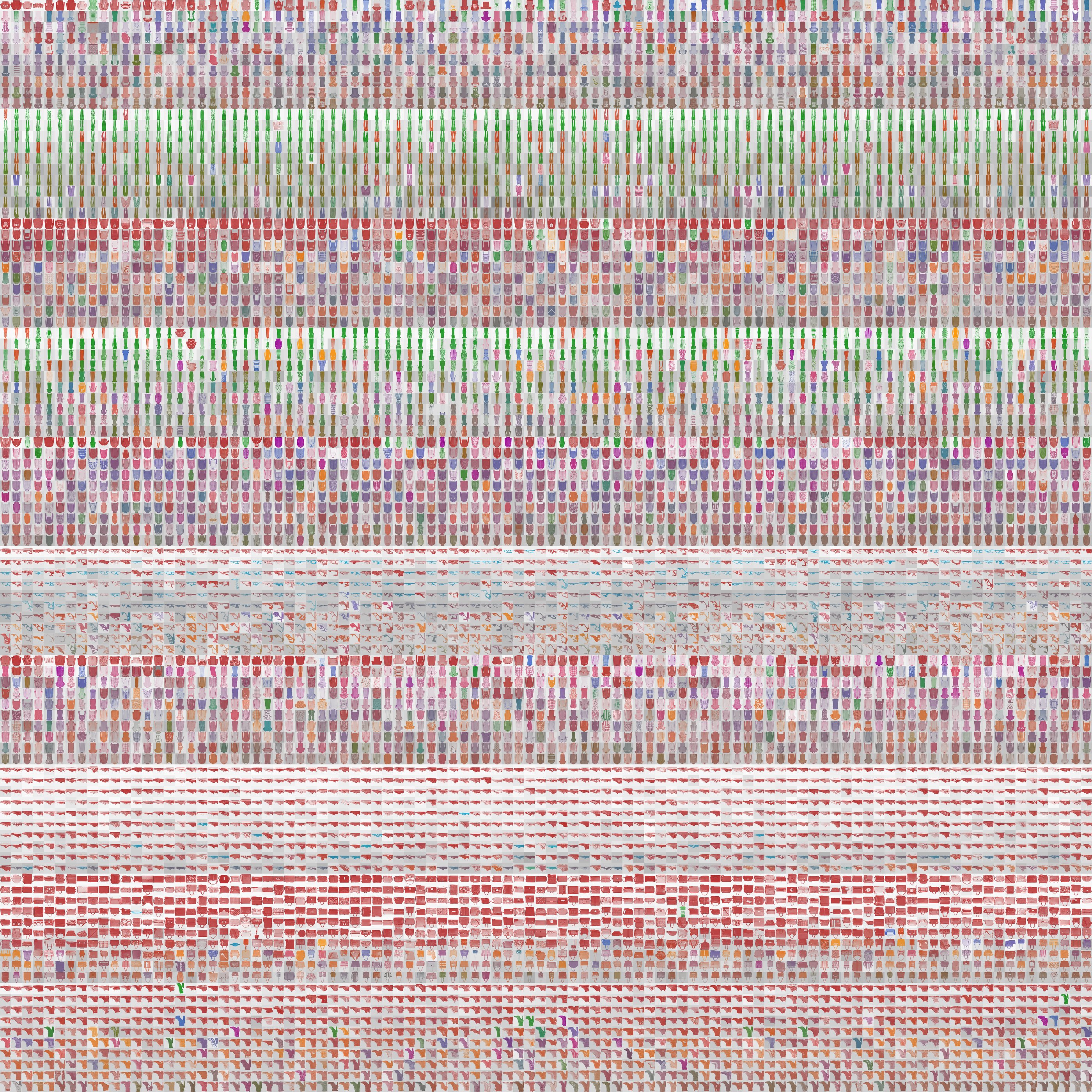}
	\caption{\label{fig:vfashion}
	The same visualization as in \Cref{fig:fashion} but for the vertically flipped FashionMNIST test set.
	}
\end{figure*}

\begin{figure*}[htb]
	\centering
	\includegraphics[width=\textwidth]{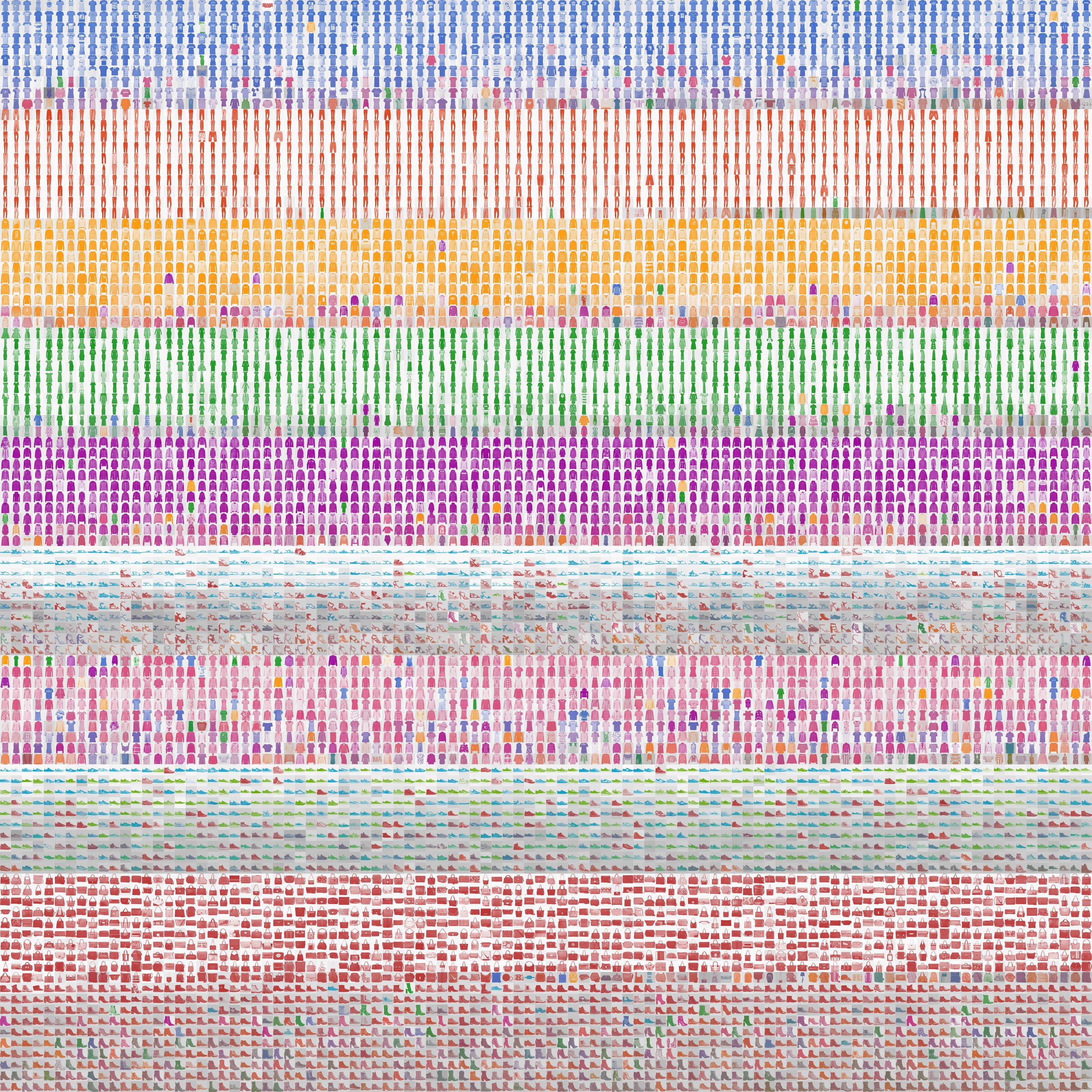}
	\caption{\label{fig:hfashion}
	The same visualization as in \Cref{fig:fashion} but for the horizontally flipped FashionMNIST test set.
	}
\end{figure*}

\Cref{fig:fashion,fig:vfashion,fig:hfashion} show the complete FashionMNIST test sets, with $p(y|x)$, $H(y|x)$,
and $R$ visualized in orthogonal manners. Comparing the Figures against one another,
you can see interesting relationships between the classes and the mirror transformations.  For instance
there is some tendency for vertically flipped pants to classify as dresses and vice versa. Under horizontal
flips most classes are well-behaved, while all of the footwear classes show marked increase in the rates.

\Cref{tab:correct-extended,tab:ood-extended} give a more detailed view of the different models
and datasets used. In these tables, we present all four different metrics: $p(y|x)$, the standard signal
used in out-of-distribution detection; $H(y|x)$, which generally outperforms $p(y|x)$ and can be used
anywhere $p(y|x)$ can be used; $p(y,z|x)$, which does well at improving false-positive-oriented metrics, like
FPR @95\% TPR and AUPR Out, but which requires a density model of the latent space, such as the one learned by
VIB; and $R$, the rate, which seems to perform well at out-of-distribution detection, and which also requires
a density model of $z$.

\Cref{tab:correct-extended} gives more data comparing the deterministic baselines with the VIB model for error
detection. Note that the deterministic baseline and VIB perform equivalently well across the board.
This is unsurprising, since if the models had a clear signal to discriminate between true and false positives
in the in-sample data, the optimization procedure should be able to find that signal and use it to directly
improve the objective function.

From \Cref{tab:ood-extended} we can draw a few early conclusions.
VIB generally dominates the deterministic baseline at T=1.
$R$ is strong at out-of-distribution detection.
$p(y,z|x)$ is strong at gross error detection (U(0,1) and MNIST FPR @95\% TPR), as well as subtle error
detection (HFlip and VFlip AUPR Out).
$p(y|x)$ and $H(y|x)$ without temperature scaling only dominate at error detection for the gross errors
(U(0,1) and MNIST AUPR Out).
However, $H(y|x)$ and $p(y,z|x)$ (not shown) are very responsive to temperature scaling,
giving substantial improvements across the board for both the deterministic model and VIB (not shown).

\InputIfFileExists{extendedtables}{}{\typeout{No file extendedtables.}}%
	

%% file: extendedtables.tex
\begin{table*}[htb]
  \centering
  \tiny
  \begin{tabular}{|c|c|c|c|c|c|c|}
    \hline
    Method &  \makecell{Accuracy} & Threshold & \makecell{FPR @95\% TPR $\downarrow$} & AUROC $\uparrow$ & \makecell{AUPR In $\uparrow$} & \makecell{AUPR Out $\uparrow$} \\
    \hhline{|=|=|=|=|=|=|=|}

    \multirow{2}{*}{\makecell{Baseline\\T=1}} & \multirow{4}{*}{92.9}
       & $p(y|x)$ &  \textbf{44.8} & \textbf{91.9} & 87.8 & 03.8 \\
     & & $H(y|x)$ &  45.3 & 91.3 & \textbf{99.3} & 03.8 \\
    \cline{0-0}
    \cline{3-7}
    \multirow{2}{*}{\makecell{Baseline\\T=100}} &
       & $p(y|x)$ &  65.2 & 87.1 & 98.9 & 03.9 \\
     & & $H(y|x)$ &  83.0 & 74.7 & 97.6 & \textbf{04.3} \\

    \hhline{|=|=|=|=|=|=|=|}

    \multirow{4}{*}{\makecell{VIB\\T=1}} & \multirow{4}{*}{92.8}
       & $p(y|x)$ &  45.6 & \textbf{\textit{91.3}} & \textbf{\textit{99.2}} & 03.8 \\
     & & $H(y|x)$ &  \textbf{44.8} & 91.2 & \textbf{\textit{99.2}} & 03.8 \\
     & & $p(y,z|x)$ &  69.3 & 83.4 & 98.3 & \textbf{\textit{04.0}} \\
     & & $R$ & 63.1 & 84.7 & 98.6 & 03.9 \\
    \hline


  \end{tabular}

  \caption{\label{tab:correct-extended}Extended results on error detection.
  Arrows denote whether higher or lower scores are better.
  \textbf{Bold} indicates the best score in that column.
  \textbf{\textit{Bold italics}} indicate the closest VIB score if a baseline score was higher.
  }
\end{table*}

\begin{table}[htb]
  \centering
  \tiny
  \setlength\tabcolsep{5pt} 
  \begin{tabular}{|c|c|c|c|c|c|c|}
  \hline
    OoD & Method & Threshold & \makecell{FPR @\\95\% TPR $\downarrow$} & \makecell{AUROC \\ $\uparrow$} & \makecell{AUPR \\ In $\uparrow$} & \makecell{AUPR \\ Out $\uparrow$} \\
    \hhline{|=|=|=|=|=|=|=|}

  \multirow{8}{*}{U(0,1)}
    & \multirow{2}{*}{\makecell{Determ.\\T=1}}
      & $p(y|x)$ &  81.7 & 86.8 & 80.2 & \textbf{33.0} \\
    & & $H(y|x)$ &  79.5 & 87.3 & 91.2 & 32.9 \\
    \cline{2-7}
    & \multirow{2}{*}{\makecell{Determ.\\T=100}}
      & $p(y|x)$ &  10.5 & 97.7 & 98.4 & 31.0 \\
    & & $H(y|x)$ &  \textbf{00.6} & \textbf{99.2} & \textbf{99.5} & 30.7 \\
    \cline{2-7}
    & \multirow{4}{*}{\makecell{VIB\\T=1}}
      & $p(y|x)$ & 54.7  & 90.0 & 90.6 & \textbf{\textit{32.0}} \\
    & & $H(y|x)$ & 43.8  & 90.8 & 91.0 & 31.8 \\
    & & $p(y,z|x)$ & 14.4 & 95.9 & 96.8 & 31.0 \\
    & & $R$ & \textbf{\textit{09.6}}  & \textbf{\textit{97.6}} & \textbf{\textit{98.0}} & 30.9 \\

  \hhline{|=|=|=|=|=|=|=|}

  \multirow{8}{*}{MNIST}
    & \multirow{2}{*}{\makecell{Determ.\\T=1}}
      & $p(y|x)$ &  82.6 & 72.9 & 64.0 & \textbf{36.2} \\
    & & $H(y|x)$ &  78.4 & 73.2 & 75.4 & 36.1 \\
    \cline{2-7}
    & \multirow{2}{*}{\makecell{Determ.\\T=100}}
      & $p(y|x)$ &  51.9 & 85.5 & 86.3 & 32.7 \\
    & & $H(y|x)$ &  41.9 & 88.9 & 90.8 & 31.8 \\
    \cline{2-7}
    & \multirow{4}{*}{\makecell{VIB\\T=1}}
      & $p(y|x)$ & 69.6  & 82.4 & 83.0 & \textbf{\textit{33.5}} \\
    & & $H(y|x)$ & 63.3  & 83.2 & 83.2 & 33.4 \\
    & & $p(y,z|x)$ &  39.7 & 86.1 & 82.6 & 32.8 \\
    & & $R$ & \textbf{27.3}  & \textbf{92.7} & \textbf{92.0} & 31.4 \\

  \hhline{|=|=|=|=|=|=|=|}

  \multirow{8}{*}{HFlip}
    & \multirow{2}{*}{\makecell{Determ.\\T=1}}
      & $p(y|x)$ &  88.0 & 65.7 & 56.1 & 38.7 \\
    & & $H(y|x)$ &  86.7 & 66.0 & 64.4 & 39.8 \\
    \cline{2-7}
    & \multirow{2}{*}{\makecell{Determ.\\T=100}}
      & $p(y|x)$ &  79.2 & \textbf{68.9} & \textbf{65.7} & 38.9 \\
    & & $H(y|x)$ &  77.1 & 67.8 & 64.7 & 39.3 \\
    \cline{2-7}
    & \multirow{4}{*}{\makecell{VIB\\T=1}}
      & $p(y|x)$ & 83.7  & 65.1 & 59.5 & 42.5 \\
    & & $H(y|x)$ & 81.1  & 65.7 & 59.8 & 42.3 \\
    & & $p(y,z|x)$ & 79.0 & 62.5 & 57.3 & \textbf{43.7} \\
    & & $R$ & \textbf{73.0}  & \textbf{\textit{68.7}} & \textbf{\textit{64.4}} & 39.3 \\

  \hhline{|=|=|=|=|=|=|=|}

  \multirow{8}{*}{VFlip}
    & \multirow{2}{*}{\makecell{Determ.\\T=1}}
      & $p(y|x)$ &  70.8 & 85.1 & 76.9 & 33.0 \\
    & & $H(y|x)$ &  64.5 & 85.6 & 87.8 & 32.8 \\
    \cline{2-7}
    & \multirow{2}{*}{\makecell{Determ.\\T=100}}
      & $p(y|x)$ &  39.4 & 91.8 & 92.6 & 31.6 \\
    & & $H(y|x)$ &  38.5 & \textbf{92.1} & \textbf{94.1} & 31.5 \\
    \cline{2-7}
    & \multirow{4}{*}{\makecell{VIB\\T=1}}
      & $p(y|x)$ & 53.7  & 84.5 & 82.5 & 32.9 \\
    & & $H(y|x)$ & 45.0  & 86.4 & 83.4 & 32.6 \\
    & & $p(y,z|x)$ & 48.1 & 83.4 & 80.7 & \textbf{33.4} \\
    & & $R$ & \textbf{34.4}  & \textbf{\textit{89.7}} & \textbf{\textit{87.0}} & 32.0 \\
    \hline
  \end{tabular}

  \caption{\label{tab:ood-extended}Extended results for out-of-distribution detection.
  Arrows denote whether higher or lower scores are better.
  \textbf{Bold} indicates the best score in that column for a particular out-of-distribution dataset.
  \textbf{\textit{Bold italics}} indicate the closest VIB score if a baseline score was higher.
  Here we can see in more detail how the four different metrics compare with each other.
  }
\end{table}